\documentclass[11pt]{article}

\usepackage[letterpaper,margin=1in,headsep=0.22in]{geometry}
\usepackage[utf8]{inputenc}
\usepackage[T1]{fontenc}
\usepackage{lmodern}

\usepackage{microtype}
\usepackage{graphicx}
\usepackage{caption}
\usepackage{booktabs}
\usepackage{adjustbox}
\usepackage{url}
\usepackage{amsmath}
\usepackage{amssymb}
\usepackage{multirow}
\usepackage[table]{xcolor}
\usepackage{float}
\usepackage[round,authoryear]{natbib}
\usepackage[hidelinks]{hyperref}
\hypersetup{
  pdftitle={Convolution for Large Language Models},
  pdfauthor={Yuchuan Tian et al.},
  pdfsubject={Technical Report}
}
\usepackage{fancyhdr}

\renewcommand{\headrulewidth}{0.4pt}
\fancypagestyle{plain}{%
  \fancyhf{}
  \fancyfoot[C]{\thepage}
  \renewcommand{\headrulewidth}{0pt}
}

\newenvironment{compacttable}[1][\linewidth]{%
  \begingroup
  \small
  \setlength{\tabcolsep}{4.2pt}%
  \begin{adjustbox}{max width=#1}%
}{%
  \end{adjustbox}%
  \endgroup
}

\title{Convolution for Large Language Models\\[-0.1em]
  \large Technical Report}

\author{%
  Yuchuan Tian\textsuperscript{1}, Yingte Shu\textsuperscript{1}, Wei He\textsuperscript{2},
  Shuo Zhang\textsuperscript{2}, Tianchen Zhao\textsuperscript{3},\\
  Chao Xu\textsuperscript{1}, Xinghao Chen\textsuperscript{2}, Yunhe Wang\textsuperscript{2},
  Hanting Chen\textsuperscript{2,\textdagger}, Yu Wang\textsuperscript{3}\\[0.45em]
  \small \textsuperscript{1}Peking University \quad
  \textsuperscript{2}Huawei Technologies \quad
  \textsuperscript{3}Tsinghua University
}
\date{July 2026}

\begin{document}
\frenchspacing

\maketitle
\begingroup
\renewcommand{\thefootnote}{\textdagger}
\footnotetext[1]{Corresponding author.}
\endgroup

\begin{abstract}

Large language models (LLMs) largely rely on Transformers, where self-attention provides global token interaction but does not explicitly encode the locality of natural language. We study whether lightweight depthwise convolutions can supply this local inductive bias without materially increasing model size. Our macro-level ablation compares convolution at 17 locations in a Qwen3 Transformer block and finds the best results when convolution is applied to the projected queries, keys, and values before attention. A subsequent micro-level study favors a residual depthwise convolution with kernel size $k=3$, without additional normalization or activation. Across Qwen3 models and several pre-training data budgets, this design improves the average accuracy on seven downstream benchmarks while adding less than $0.01\%$ parameters. A representation-level case study further suggests that the convolution makes repeated token IDs more sensitive to their immediate context. These results support depthwise convolution as a lightweight complement to self-attention for modeling short-range token interactions.

\end{abstract}

\section{Introduction}

Large language models (LLMs), including DeepSeek \citep{liu2024deepseek} and Llama \citep{grattafiori2024llama}, are predominantly built by scaling Transformer architectures. Their basic structure remains a stack of self-attention layers for token interaction and feed-forward networks (FFNs) for token-wise feature transformation. Variants such as Multi-Head Latent Attention (MLA) and Grouped-Query Attention (GQA) improve the efficiency of attention, but local composition is still learned implicitly rather than encoded directly in the architecture.

\begin{figure}[t]
    \centering
    \includegraphics[width=0.92\textwidth]{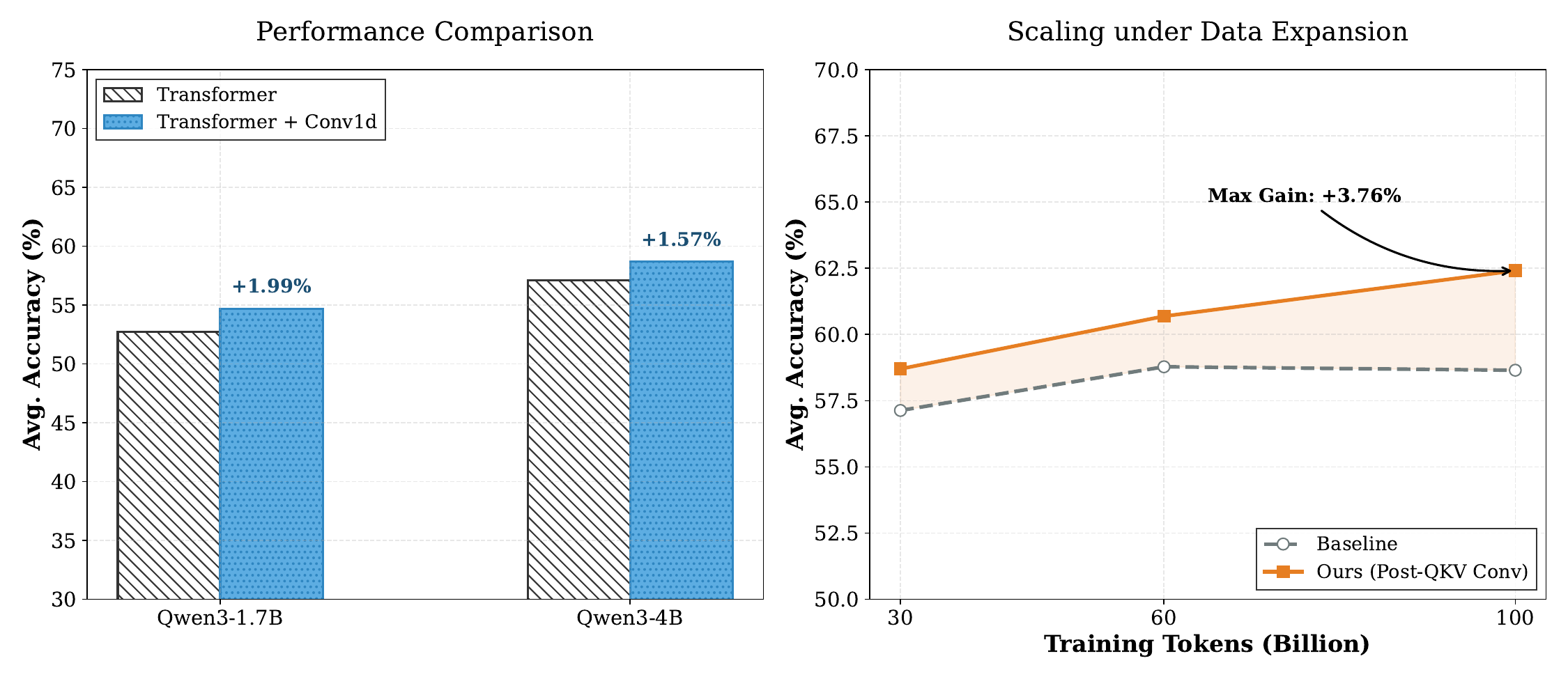}
    \caption{Summary of the downstream evaluation. Left: average accuracy at the 30B-token training budget for the evaluated Qwen3 model sizes. Right: average accuracy of Qwen3-4B at three training-data budgets. The post-QKV depthwise convolution adds less than $0.01\%$ parameters.}
    \label{fig:performance_scaling}
\end{figure}

Natural language contains substantial short-range structure: adjacent tokens form compounds, phrases, and other units that contribute to sentence-level meaning. For example, in ``neural network optimization,'' the tokens ``neural'' and ``network'' form a familiar local unit. Transformers can learn such patterns through self-attention, but the standard architecture provides no explicit preference for them. This motivates adding a lightweight operator whose receptive field is local by construction.

We therefore revisit one-dimensional convolution as a way to add locality to modern LLMs. Unlike hybrid architectures that use convolution as a separate block, such as Conformer \citep{gulati2020conformer}, we use it as a small refinement module within the existing Transformer data path. Our working hypothesis is that aggregating nearby features before attention can provide locally contextualized representations for subsequent long-range interaction.

We organize the study from macro-level placement to micro-level module design. At the macro level, we insert depthwise Conv1D at 17 locations in a Qwen3 Transformer block. The strongest configuration applies convolution after the query, key, and value projections and before attention. At the micro level, we compare residual connections, normalization, kernel size, initialization, activation functions, and reparameterization. Depthwise convolution operates independently on each channel, keeping the added parameter count linear in the hidden dimension.

Across the evaluated model sizes and training-data budgets, the resulting model improves average downstream accuracy over its matched Transformer baseline. Figure~\ref{fig:performance_scaling} summarizes these results. We also examine a WSC example to assess how the convolution changes the representations of repeated token IDs in different local contexts.

The main contributions are as follows:

\begin{enumerate}
    \item We compare 17 convolution locations in a Qwen3 Transformer block and find that post-QKV convolution performs best in our controlled placement study.
    \item We identify a compact module design: a residual depthwise Conv1D with kernel size 3 and no added normalization or activation.
    \item We evaluate the design across Qwen3 model sizes and training-data budgets, observing higher average accuracy with less than $0.01\%$ additional parameters.
\end{enumerate}

\section{Related Work}

Convolution has often been combined with Transformer architectures to supplement the global receptive field of self-attention with local structure. We review representative approaches in vision and language modeling.

\subsection{Convolution in Transformers}

In computer vision, convolution is incorporated at several points in a Transformer. LeViT \citep{graham2021levit} and CoAtNet \citep{dai2021coatnet} use convolutional embeddings, downsampling, or convolutional blocks alongside attention. CvT \citep{wu2021cvt} applies convolutional projections before producing attention inputs, whereas ConViT \citep{d2021convit} introduces a gated positional bias toward local context. TiC \citep{zhang2023tic} develops a multi-head self-attention convolution as an alternative to global attention. Convolution also appears within the FFN in PVTv2 \citep{wang2022pvt}, SRFormer \citep{zhou2023srformer}, and CeiT \citep{yuan2021incorporating}. Conformer \citep{gulati2020conformer} places a convolution module between the FFN and attention sublayers, and depthwise convolution has also been used to bypass selected Transformer blocks \citep{zhang2025depth}. In language modeling, ConvBERT \citep{jiang2020convbert} replaces a subset of attention heads with span-based dynamic convolution.

\subsection{Locality in Language Models}

Several language-model architectures capture local or sequential structure through convolution-like operations. RWKV \citep{peng2023rwkv} combines linear attention with time decay, and the Gated Delta Network \citep{yang2024gated} uses recurrent state updates. Multi-Token Attention \citep{golovneva2025multi} allows queries to attend to contiguous token blocks. Zoology \citep{arora2023zoology} and work on simple linear attention \citep{arora2024simple} study the role of local modeling in recall and pattern-matching tasks. Other approaches represent recurring phrases with scalable lookup tables \citep{cheng2026conditional} or use parameterized long convolutions to capture dependencies at multiple scales \citep{poli2023hyena}. These models differ in mechanism and scope, but they all provide alternatives to unrestricted token-to-token attention for modeling sequential structure.

Our focus is narrower: we keep the Transformer backbone fixed and compare where and how a lightweight depthwise convolution can be inserted. This controlled design study separates the effects of placement, residual structure, normalization, kernel size, initialization, activation, and reparameterization while preserving the backbone's global attention mechanism.

\section{Designing Convolutional Augmentation}

\subsection{A Lightweight Local Operator}
Our design objective is to add a local inductive bias with little parameter overhead and minimal disruption to the Transformer block. We therefore use \textbf{depthwise Conv1D} as the basic operator and do not include dense cross-channel mixing within the convolutional layer.

For a standard 1D convolution with kernel size $k$, the parameter count scales as $\mathcal{O}(k d_{\mathrm{model}}^2)$. Qwen3-1.7B has $L=28$ layers and hidden size $d_{\mathrm{model}}=2048$. Adding a standard convolution with $k=2$ to every layer would therefore introduce approximately $2 \times 2048^2 \times 28 \approx 234.88$ million parameters, or about $13.5\%$ of the baseline parameter count.

Depthwise Conv1D instead operates within each channel, reducing the parameter complexity to $\mathcal{O}(k d_{\mathrm{model}})$. Under the same calculation, it adds approximately $2 \times 2048 \times 28 \approx 0.11$ million parameters, corresponding to about \textbf{0.006\%} of the baseline. Channel mixing remains the responsibility of the existing projection and FFN layers.

\subsection{Comparing Convolution Locations}

\begin{table}[H]
\centering
\caption{Candidate Conv1D locations in a Qwen3 Transformer block. We report the convolution input width, average loss, WikiText-103 perplexity, and total parameter count in millions.}
\label{tab:conv_positions}
\begin{compacttable}[\textwidth]
\begin{tabular}{cccccc}
\toprule
\textbf{ID} & \textbf{Position} & \textbf{Width} & \textbf{Mean loss} & \textbf{Perplexity} & \textbf{Params (M)}\\
\midrule
No & No convolution added & -- & 2.5144 & 13.42 & 1720.57\\
\midrule
P1  & Pre-attention residual branch input & $H$ & 2.5056 & 13.27 & 1721.03 \\
P2  & Post-attention residual branch, pre-attention normalization & $H$ & 2.5000 & 13.18 & 1720.86 \\
P3  & Inside attention residual path (pre- or post-split) &  $H$ & 2.5554 & 14.31 & 1720.86 \\
P4  & Post-attention RMSNorm, pre-QKV linear projections &  $H$ & 2.4906 & 13.06 & 1720.86 \\
\textbf{P5}  & \textbf{On concatenated QKV projection outputs} &  $H_{\text{q}}+2H_{\text{kv}}$ & \textbf{2.4818} & \textbf{12.85} & \textbf{1721.15} \\
P6  & On projected Query representations only &  $H_{\text{q}}$ & 2.5115 & 13.38 & 1720.80 \\
P7  & On projected Key representations only &  $H_{\text{kv}}$ & 2.5085 & 13.29 & 1720.72\\
P8  & On projected Value representations only &  $H_{\text{kv}}$ & 2.4962 & 13.21 & 1720.72 \\
P9  & Between attention score computation and output projection &  $H$  & 2.4986 & 13.40 & 1720.86  \\
P10 & Post-attention submodule, pre-residual addition &  $H$  & 2.5020 & 13.31 & 1720.86  \\
P11 & Between attention residual output and FFN residual branch &  $H$  & 2.5002 & 13.36 & 1720.86 \\
P12 & Inside FFN (SwiGLU) residual path (pre- or post-split) & $H$  & 2.4884 & 13.02 & 1720.86  \\
P13 & Post-FFN residual branch, pre-SwiGLU body & $H$  & 2.5028 & 13.38 & 1720.86 \\
P14 & Post-FFN RMSNorm, pre-gating and up-projection & $H$  & 2.5069 & 13.24 & 1720.86  \\
P15 & On joint output of FFN gating and up-projection &  $2H_{\text{ffn}}$  & 2.5044 & 13.29 & 1722.30  \\
P16 & Post element-wise gating multiplication, pre-down-projection &  $H_{\text{ffn}}$ &   2.4872 & 13.30 & 1721.44\\
P17 & Post FFN down-projection & $H$  & 2.4877 & 13.12 & 1720.86 \\
\bottomrule
\end{tabular}
\end{compacttable}
\end{table}

\begin{figure}[t]
    \centering
    \includegraphics[width=0.66\textwidth]{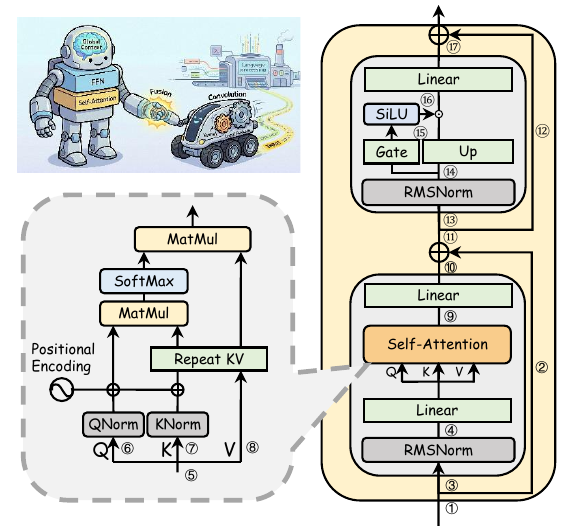}
    \caption{A Qwen3 Transformer block and the 17 candidate locations for a Conv1D module.}
    \label{fig:block}
\end{figure}

Figure~\ref{fig:block} summarizes a Qwen3 Transformer block \citep{yang2025qwen3}. The input $\mathbf{u}$ is first normalized with RMSNorm to produce $\mathbf{h}$. Linear projections then produce the queries, keys, and values; QK normalization and rotary positional embeddings (RoPE) are applied before grouped-query attention. The attention output is projected and added to $\mathbf{u}$, producing the intermediate state $\mathbf{x}$. A second RMSNorm and a SwiGLU FFN follow. In the FFN, separate gate and up projections are combined by applying $\mathrm{Swish}$ to the gate branch and multiplying it element-wise by the up branch. A down projection maps the result back to the model dimension before the second residual addition.

Table~\ref{tab:conv_positions} defines the 17 candidate locations. We use Qwen3-1.7B as the backbone, add one Conv1D module at each location, and train each model from scratch on FineWeb-100B \citep{penedo2024fineweb}. We report the mean training loss over the final 10,000 iterations, perplexity on WikiText-103 \citep{merity2018scalable}, and total parameter count. For discussion, we group the locations by whether they operate directly on the inputs or outputs of attention.

The first group contains locations before attention, after attention, or in the FFN pathway (P1--P4 and P10--P17). At these locations, convolution acts on representations either before they enter attention or after global mixing. Most configurations improve at least one of loss or perplexity, but the gains vary substantially; P3 performs worse than the baseline on both metrics.

The second group, P5--P9, acts directly on attention components. P6--P8 convolve only the query, key, or value representation, whereas P5 jointly processes the concatenated QKV projections. P9 applies convolution after attention scores have been used for aggregation. Among these choices, P5 obtains the lowest average loss (2.4818) and perplexity (12.85). One possible interpretation is that P5 supplies local context to all three attention inputs before global aggregation; the current ablation establishes the empirical ordering but does not isolate this mechanism.

We therefore use P5, the post-QKV location, in the remaining design experiments.

\subsection{Comparing Convolution Module Designs}

\begin{figure}[H]
    \centering
    \includegraphics[width=0.68\textwidth]{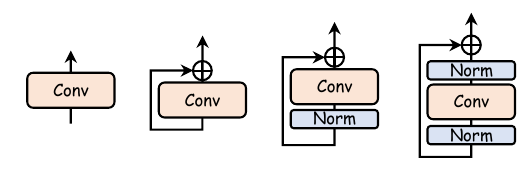}
    \caption{The four Conv1D module designs. From left to right: plain convolution, convolution with a shortcut, convolution with pre-normalization, and convolution with sandwich normalization.}
    \label{fig:transformerhow}
\end{figure}

\paragraph{Module design.} We first compare four internal configurations of the depthwise convolution module (Figure~\ref{fig:transformerhow}):
(1) \textbf{plain convolution}, $Y = \operatorname{Conv}(X)$;
(2) \textbf{convolution with a shortcut}, $Y = X + \operatorname{Conv}(X)$;
(3) \textbf{convolution with pre-normalization}, $Y = X + \operatorname{Conv}(\operatorname{Norm}(X))$; and
(4) \textbf{convolution with sandwich normalization}, $Y = X + \operatorname{Norm}(\operatorname{Conv}(\operatorname{Norm}(X)))$.
% !!!NEED UPDATE!

% Table generated by Excel2LaTeX from sheet 'Sheet1'
\begin{table}[htbp]
  \centering
  \caption{Effect of the Conv1D module design on Qwen3-1.7B.}
  \label{tab:micro}
  \begin{compacttable}
        \begin{tabular}{cccc}
        \toprule
        \textbf{Configuration} & \textbf{Mean loss} & \textbf{Perplexity} & \textbf{Params (M)} \\
        \midrule
        No Conv1D & 2.5144 & 13.42 & 1720.57 \\
        \midrule
        Convolution & 2.4931 & 13.05 & 1721.15 \\
        \textbf{Conv + Shortcut} & \textbf{2.4795} & \textbf{12.79} & \textbf{1721.03} \\
        Pre-Norm & 2.4876 & 12.94 & 1721.03 \\
        Sandwich Norm & 2.5110 & 13.33 & 1721.26 \\
        \bottomrule
        \end{tabular}%
    \end{compacttable}
\end{table}%

All four convolutional variants improve average loss and perplexity over the ``No Conv1D'' baseline in Table~\ref{tab:micro}.

The \textbf{Conv + Shortcut} configuration gives the lowest reported perplexity, 12.79. The shortcut preserves the projected QKV features and lets the convolution learn a residual local correction.

Adding normalization does not improve this result: pre-normalization yields 12.94 perplexity, and sandwich normalization yields 13.33. Because the block already normalizes its input, an additional normalization at the post-QKV stage may alter the feature scale used by attention. We therefore retain the simpler \textbf{Conv + Shortcut} design for the remaining experiments.

\subsection{Configuration Ablations}

\begin{table}[htbp]
  \centering
  \caption{Effect of kernel size on Qwen3-1.7B.}
  \label{tab:kernel}
      \begin{compacttable}
        \begin{tabular}{cccc}
        \toprule
        \textbf{Kernel size} & \textbf{Mean loss} & \textbf{Perplexity} & \textbf{Params (M)} \\
        \midrule
        No Conv1D & 2.5144 & 13.42 & 1720.57 \\
        \midrule
        $k=2$ & 2.4894 & 12.99 & 1720.92 \\
        \textbf{$k=3$} & \textbf{2.4795} & \textbf{12.79} & \textbf{1721.03} \\
        $k=4$ & 2.4881 & 13.13 & 1721.15 \\
        \bottomrule
        \end{tabular}%
      \end{compacttable}
\end{table}%

\paragraph{Kernel size.} We compare $k \in \{2,3,4\}$ for the depthwise convolution at the post-QKV location. As shown in Table~\ref{tab:kernel}, all three settings improve over the baseline, and $k=3$ gives the lowest reported perplexity, 12.79. Increasing the kernel size to $k=4$ does not yield a further improvement. This result is consistent with a compact kernel being sufficient for short-range composition, although the ablation does not identify why $k=3$ performs best. We use $k=3$ in the remaining experiments.

\paragraph{Initialization.}

% Table generated by Excel2LaTeX from sheet 'Sheet1'
\begin{table}[htbp]
  \centering
  \caption{Effect of Conv1D initialization on Qwen3-1.7B. ``Random'' denotes sampling the weight or bias from $\mathcal{N}(0,0.02)$.}
  \label{tab:addinit}
  \begin{compacttable}
        \begin{tabular}{ccccc}
        \toprule
    \textbf{Weight} & \textbf{Bias} & \textbf{Mean loss} & \textbf{Perplexity} & \textbf{Params (M)} \\
        \midrule
    zero  & N/A   & 3.0065 & 61.52 & 1720.92 \\
    zero  & zero  & 2.5056 & 13.27 & 1721.03 \\
    zero  & random & 2.4908 & 12.85 & 1721.03 \\
    \textbf{random} & \textbf{random} & \textbf{2.4795} & \textbf{12.79} & \textbf{1721.03} \\
        \bottomrule
    \end{tabular}%
    \end{compacttable}
\end{table}%

Table~\ref{tab:addinit} compares four initialization settings. Random initialization of both weights and biases gives the lowest reported loss and perplexity. The bias-free, zero-weight setting performs substantially worse, whereas the zero-weight setting with a zero or random bias does not show the same degradation. We use random initialization for both weights and biases in the remaining experiments.

\subsection{Additional Architectural Variants}

After selecting the location, module design, kernel size, and initialization, we test three additional choices: a second convolution, nonlinear activation, and multi-branch reparameterization. None improves both loss and perplexity over the selected configuration.

\begin{table}[htbp]
  \centering
  \caption{Results for Qwen3-1.7B with Conv1D at P5, P17, or both locations.}
  \label{tab:addmulti}
  \begin{compacttable}
    \begin{tabular}{cccc}
            \toprule
        \textbf{Position} & \textbf{Mean loss} & \textbf{Perplexity} & \textbf{Params (M)} \\
        \midrule
    \textbf{P5}  & \textbf{2.4818} & \textbf{12.85} & \textbf{1721.15} \\
    P17 & 2.4877 & 13.12 & 1720.86 \\
    P5+P17 & 2.4836 & 13.04 & 1721.44 \\
    \bottomrule
    \end{tabular}%
  \end{compacttable}
\end{table}%

\paragraph{Multiple convolution modules.}
We add a second Conv1D module at P17 to the selected P5 configuration. As shown in Table~\ref{tab:addmulti}, this variant slightly increases both loss and perplexity. The result may reflect redundant local mixing after attention and the FFN, but the ablation establishes only that adding P17 is not beneficial in this setting. We therefore retain a single convolution at P5.

\begin{table}[htbp]
  \centering
  \caption{Effect of activation functions in the Qwen3-1.7B Conv1D module.}
  \label{tab:addact}
  \begin{compacttable}
    \begin{tabular}{cccc}
            \toprule
        \textbf{Configuration} & \textbf{Mean loss} & \textbf{Perplexity} & \textbf{Params (M)} \\
        \midrule
    \textbf{No activation} & \textbf{2.4795} & \textbf{12.79} & \textbf{1721.03} \\
    SiLU  & 2.4962 & 13.06 & 1721.03 \\
    LeakyReLU & 2.4892 & 12.96 & 1721.03 \\
    sigmoid & 2.4889 & 12.94 & 1721.03 \\
    \bottomrule
    \end{tabular}%
  \end{compacttable}
\end{table}%

\paragraph{Activation function.}
We test SiLU, LeakyReLU, and sigmoid activations after Conv1D. Each activation increases both loss and perplexity relative to the linear configuration in Table~\ref{tab:addact}. We therefore use a linear convolution without an activation function.

\begin{table}[htbp]
  \centering
  \caption{Effect of convolutional reparameterization in Qwen3-1.7B.}
  \label{tab:addrep}
  \begin{compacttable}
    \begin{tabular}{cccc}
        \toprule
        \textbf{Configuration} & \textbf{Mean loss} & \textbf{Perplexity} & \textbf{Params (M)} \\
        \midrule
    \textbf{No reparameterization} & \textbf{2.4795} & \textbf{12.79} & \textbf{1721.03} \\
    Kernel 1 branch & 2.5029 & 13.28 & 1721.26 \\
    Kernel 1 and 2 branches & 2.5048 & 13.28 & 1721.61 \\
    \bottomrule
    \end{tabular}%
  \end{compacttable}
\end{table}%

\paragraph{Reparameterization.}
We also train multi-branch Conv1D modules with different kernel sizes and merge the branches into an equivalent convolution for inference. Both reparameterized variants perform worse than the single-branch configuration in Table~\ref{tab:addrep}. One possible explanation is that the branches mix local patterns over different spans, although this ablation does not isolate the cause. We therefore use a single compact Conv1D module.

\section{Representation Analysis}

\begin{figure}[H]
    \centering
    \includegraphics[width=0.96\textwidth]{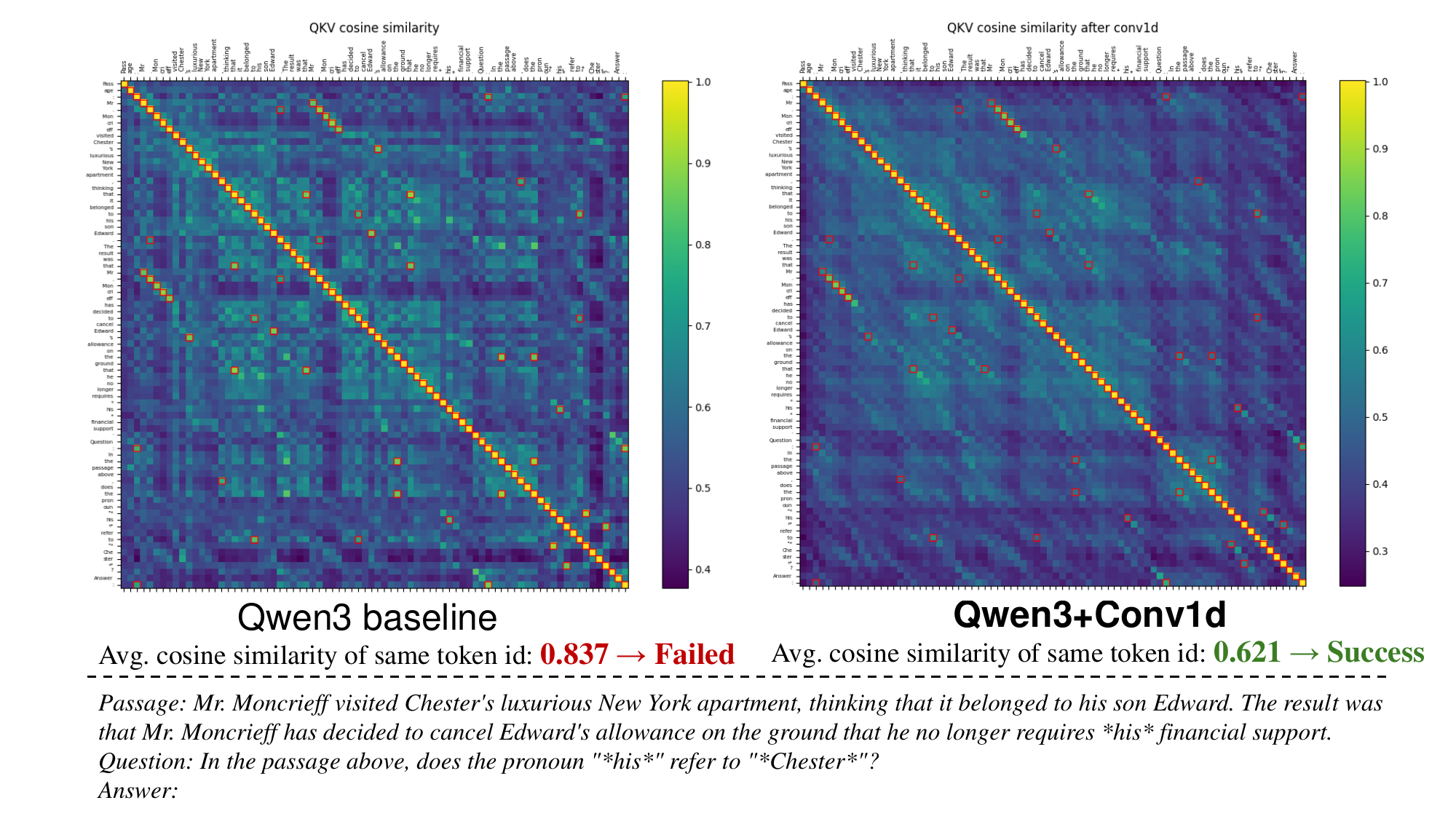}
    \caption{\textbf{Cosine-similarity heatmaps of QKV representations from the Qwen3 baseline and Qwen3+Conv1D in one WSC example.} Red boxes mark repeated token IDs at different positions. In this example, the Conv1D model assigns lower similarity to these occurrences and produces the correct prediction.}
    \label{fig:cossim}
\end{figure}

We use one WSC example to examine whether Conv1D changes the representations of repeated token IDs that occur in different local contexts. RoPE encodes token position, while the neighboring tokens may help determine the interpretation of each occurrence.

For the same prompt, we extract the QKV projection outputs from the Qwen3 baseline and its Conv1D-augmented counterpart and compute pairwise cosine similarities across positions. For repeated token IDs, lower similarity is consistent with greater contextual differentiation, although cosine similarity alone does not establish a semantic interpretation. The example is drawn from WSC \citep{levesque2012winograd}.

As shown in Figure~\ref{fig:cossim}, the highlighted repeated-token occurrences have lower cosine similarity in the Conv1D model than in the baseline. This case shows that the convolution changes QKV representations according to nearby context and is consistent with the proposed role of the module. Establishing a general mechanism would require an aggregate analysis over many examples and suitable controls.

\section{Experiments}

\begin{table}[H]
\centering
\caption{\textbf{Main results.} Accuracy on seven benchmarks for Qwen3 backbones trained with different token budgets, with and without Conv1D.}
\label{tab:main}
\begin{compacttable}[\textwidth]
\begin{tabular}{ccccccccccc}
\toprule
\textbf{Backbone} & \textbf{Data} & \textbf{Conv1D} & \textbf{ARC-C} & \textbf{ARC-E} & \textbf{BoolQ} & \textbf{HellaSwag} & \textbf{PIQA} & \textbf{WinoGrande} & \textbf{WSC} & \textbf{Average} \\

\midrule
\multirow{2}{*}{Qwen3-1.7B} & \multirow{2}{*}{30B} & No & 32.00 & 69.23 & 53.27 & 40.24 & 70.89 & 55.25 & 48.08 & 52.71 \\
                           &                        & \cellcolor{gray!20}Yes & \cellcolor{gray!20}33.28 & \cellcolor{gray!20}69.99 & \cellcolor{gray!20}55.50 & \cellcolor{gray!20}41.29 & \cellcolor{gray!20}71.11 & \cellcolor{gray!20}55.96 & \cellcolor{gray!20}55.77 & \cellcolor{gray!20}54.70 \\
\midrule
\multirow{2}{*}{Qwen3-4B}   & \multirow{2}{*}{30B} & No & 37.46 & 72.01 & 63.64 & 44.45 & 72.52 & 59.83 & 50.00 & 57.13 \\
                           &                        & \cellcolor{gray!20}Yes & \cellcolor{gray!20}38.05 & \cellcolor{gray!20}73.32 & \cellcolor{gray!20}60.52 & \cellcolor{gray!20}44.06 & \cellcolor{gray!20}72.69 & \cellcolor{gray!20}58.80 & \cellcolor{gray!20}63.46 & \cellcolor{gray!20}58.70 \\
\midrule
\multirow{2}{*}{Qwen3-4B}   & \multirow{2}{*}{60B} & No & 37.97 & 72.85 & 65.60 & 45.53 & 73.61 & 61.09 & 54.81	&58.78
 \\
                           &                        & \cellcolor{gray!20}Yes & \cellcolor{gray!20}39.42 & \cellcolor{gray!20}73.44 & \cellcolor{gray!20}65.05 & \cellcolor{gray!20}46.26 & \cellcolor{gray!20}73.18 & \cellcolor{gray!20}64.01 & \cellcolor{gray!20}63.46 & \cellcolor{gray!20}60.69 \\
\midrule
\multirow{2}{*}{Qwen3-4B}   & \multirow{2}{*}{100B} & No & 42.66 & 77.31 & 64.56 & 49.57 & 75.46 & 64.48 & 36.54 & 58.65 \\
                           &                         & \cellcolor{gray!20}Yes & \cellcolor{gray!20}44.03 & \cellcolor{gray!20}78.07 & \cellcolor{gray!20}67.95 & \cellcolor{gray!20}49.94 & \cellcolor{gray!20}76.33 & \cellcolor{gray!20}65.75 & \cellcolor{gray!20}54.81 & \cellcolor{gray!20}62.41 \\
\bottomrule
\end{tabular}
\end{compacttable}
\end{table}

\subsection{Experimental Setup}

\paragraph{Causal implementation and decoding.} To preserve autoregressive causality, we use a causal depthwise convolution. For kernel size $k$, left padding by $k-1$ positions ensures that the output at time $t$ depends only on the current token and the preceding $k-1$ tokens.

During incremental decoding, we maintain a convolution cache containing the features of the most recent $k-1$ tokens. For each new token, the convolution uses this cache and the current input; the cache is then updated as a sliding window. The convolutional work per decoded token therefore does not grow with the decoded context length.

\paragraph{Models and data.}

We pretrain Qwen3-1.7B and Qwen3-4B from scratch. Both backbones use grouped-query attention, SwiGLU activations, and rotary positional embeddings. The experiments use FineWeb-100B \citep{penedo2024fineweb} and its 30B- and 60B-token subsets.

\paragraph{Hyperparameters.}
We set the training hyperparameters based on Step Law~\citep{steplaw}, with adjustments for optimization stability. The peak learning rates are $1.0 \times 10^{-3}$ for Qwen3-1.7B and $6.0 \times 10^{-4}$ for Qwen3-4B. The learning rate is annealed after reaching its peak. All runs use a context length of 4096 tokens and a global batch size of 256 sequences, corresponding to 1,048,576 tokens per optimization step.

\paragraph{Evaluation.}
We evaluate the models on seven benchmarks: ARC-Challenge and ARC-Easy \citep{clark2018think}, BoolQ \citep{clark2019boolq}, HellaSwag \citep{zellers2019hellaswag}, PIQA \citep{bisk2020piqa}, WinoGrande \citep{sakaguchi2021winogrande}, and WSC \citep{levesque2012winograd}.

Table~\ref{tab:main} summarizes the results across backbone sizes and training-token budgets. Conv1D improves the average score in each reported setting: by 1.99 points for Qwen3-1.7B trained on 30B tokens, and by 1.57, 1.91, and 3.76 points for Qwen3-4B trained on 30B, 60B, and 100B tokens, respectively. The largest individual gains occur on WSC.

Although the average improves in every setting, several individual benchmark scores decrease. The larger average gain at 100B tokens suggests that the benefit may increase with the data budget for Qwen3-4B, but additional training budgets and repeated runs would be needed to establish a general scaling trend.

\section{Conclusion}

We study depthwise Conv1D as a lightweight local inductive bias for Transformer-based language models. Among the 17 insertion points tested, applying convolution to the QKV projections before attention gives the lowest loss and perplexity. The subsequent ablations favor a residual convolution with kernel size $k=3$, random initialization, and no additional normalization or activation. Across the evaluated Qwen3 model sizes and training-token budgets, this configuration improves average benchmark accuracy with a small increase in parameter count. These results indicate that local convolution can complement self-attention by incorporating neighboring-token information before global interaction. The evidence is limited to the evaluated Qwen3 configurations, and broader comparisons are needed to determine how well the result transfers to other model families and training settings.

% \newpage

\bibliographystyle{plainnat}
\bibliography{example_paper}

%%%%%%%%%%%%%%%%%%%%%%%%%%%%%%%%%%%%%%%%%%%%%%%%%%%%%%%%%%%%%%%%%%%%%%%%%%%%%%%
%%%%%%%%%%%%%%%%%%%%%%%%%%%%%%%%%%%%%%%%%%%%%%%%%%%%%%%%%%%%%%%%%%%%%%%%%%%%%%%
% APPENDIX
%%%%%%%%%%%%%%%%%%%%%%%%%%%%%%%%%%%%%%%%%%%%%%%%%%%%%%%%%%%%%%%%%%%%%%%%%%%%%%%
%%%%%%%%%%%%%%%%%%%%%%%%%%%%%%%%%%%%%%%%%%%%%%%%%%%%%%%%%%%%%%%%%%%%%%%%%%%%%%%
% \newpage
% \appendix
% \onecolumn
% \section{Appendix A}

% appendix appendix

%%%%%%%%%%%%%%%%%%%%%%%%%%%%%%%%%%%%%%%%%%%%%%%%%%%%%%%%%%%%%%%%%%%%%%%%%%%%%%%
%%%%%%%%%%%%%%%%%%%%%%%%%%%%%%%%%%%%%%%%%%%%%%%%%%%%%%%%%%%%%%%%%%%%%%%%%%%%%%%

\end{document}